\documentclass[10pt,twocolumn,letterpaper]{article}

\usepackage{cvpr}
\usepackage{times}
\usepackage{epsfig}
\usepackage{graphicx}
\usepackage{amsmath}
\usepackage{amssymb}

\usepackage[caption=false]{subfig}
\usepackage[ruled,vlined]{algorithm2e}
\usepackage{amsfonts}
\usepackage{amssymb}
\usepackage{gensymb}
\usepackage{multirow}
\usepackage{tabularx}

\usepackage[breaklinks=true,bookmarks=false]{hyperref}

\cvprfinalcopy 

\def\cvprPaperID{****} 

\begin{document}

\title{Matching Neuromorphic Events and Color Images via Adversarial Learning}

\author{Fang Xu, Shijie Lin, Wen Yang, Lei Yu, Dengxin Dai, Gui-song Xia\\
}

\maketitle

\begin{abstract}
	The event camera has appealing properties: high dynamic range, low latency, low power consumption and low memory usage, and thus provides complementariness to conventional frame-based cameras. It only captures the dynamics of a scene and is able to capture almost “continuous” motion. However, different from frame-based camera that reflects the whole appearance as scenes are, the event camera casts away the detailed characteristics of objects, such as texture and color. To take advantages of both modalities,  the event camera and frame-based camera are combined together for various machine vision tasks. Then the cross-modal matching between neuromorphic events and color images plays a vital and essential role. In this paper, we propose the Event-Based Image Retrieval (EBIR) problem to exploit the cross-modal matching task. Given an event stream depicting a particular object as query, the aim is to retrieve color images containing the same object. This problem is challenging because there exists a large modality gap between neuromorphic events and color images. We address the EBIR problem by proposing neuromorphic Events-Color image Feature Learning (ECFL). Particularly, the adversarial learning is employed to jointly model neuromorphic events and color images into a common embedding space. We also contribute to the community N-UKbench and EC180 dataset to promote the development of EBIR problem. Extensive experiments on our datasets show that the proposed method is superior in learning effective modality-invariant representation to link two different modalities.
\end{abstract}

\section{Introduction}
\begin{figure}[t]
	\centering
	\includegraphics[width=\columnwidth]{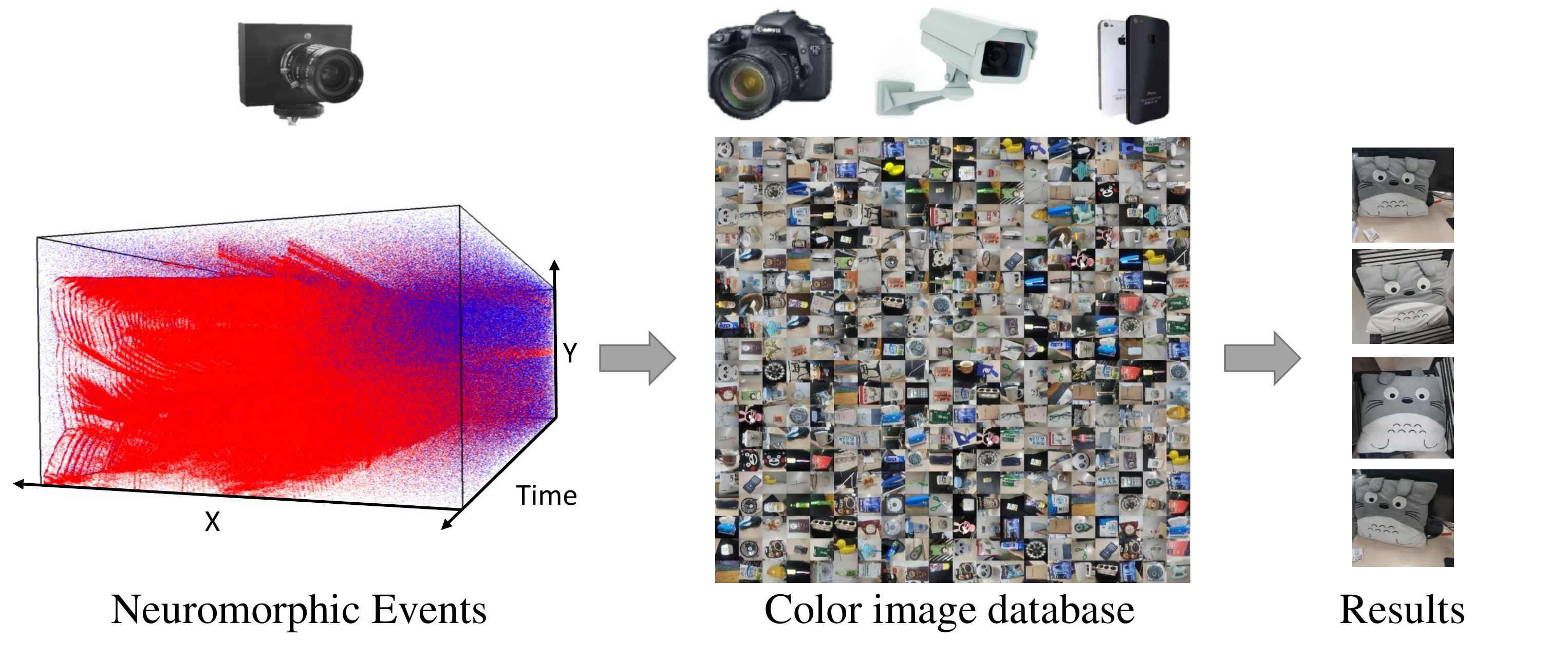} 
	\caption{The illustration of Event-Based Image Retrieval (EBIR) task. In this task, we try to use an event stream as query to retrieve the color images describing the same object.}
	\label{fig:mission}
\end{figure}

The event camera is bio-inspired, event-driven, time-based neuromorphic vision sensor, which senses the world on a radically different principle~\cite{2008dvs}. Instead of capturing static images of the scene at a constant acquisition rate like the conventional frame-based camera, the event camera works asynchronously to measure the brightness changes and reports an event once the change exceeds a threshold. It casts away the concepts like exposure time and frame that have dominated the computer vision community over decades, thus  
enabling to provide compensation to the shortcomings of the frame-based camera.  
For example, the frame-based camera represents motion by capturing a series of still frames, which results in the loss of information between frames. Instead, the event camera represents motion by capturing an event stream, which enables to capture almost “continuous” motion in frame-free mode.  Beyond that, the event camera only records what changes, which drastically reduces power, data storage and computational requirements. Such unique advantages drive the event-based vision to disrupt the current technology in fields such as automotive vehicles, security and surveillance~\cite{gallego2019event, pikatkowska2012spatiotemporal, Wang_2019_CVPR, zanardi2019cross}.

However, unlike the frame-based camera encoding illumination intensity to depict the object/scene as it is in real world, the event camera encodes only brightness changes. It makes the detailed characteristics of object/scene that recorded in the event camera, such as texture information and color information, still have to be obtained from the traditional frame-based camera. As a consequence, we are encountering the situation where the event camera and frame-based camera are combined together for various machine vision tasks, such as a distributed computer machine vision system composed of these two sensors for surveillance and monitoring. Then how to match the interested object from these two modalities of data becomes critical.
In this paper, to study the cross-modal matching between neuromorphic events and color images, the problem of Event-Based Image Retrieval (EBIR) is proposed. Given an event stream as query, the aim is to retrieve the color images describing the same objects from an image database, as shown in Fig.~\ref{fig:mission}. This is an extremely challenging problem due to the following aspects: (\romannumeral1) The output of event camera is not a sequence of frames but a sparse, asynchronous stream of events, which makes the commonly used methods for conventional frame-based camera are not directly applicable. (\romannumeral2) A large modality gap exists between neuromorphic events and color images due to the inconsistent distributions, which arises many difficulties to infer the joint distribution or learn the common representations. (\romannumeral3) A fit-for-purpose dataset is lacking. None of existed datasets can provide an evaluation for EBIR task. 

We address all above challenges by proposing an EBIR framework that we refer to as neuromorphic Events-Color image Feature Learning (ECFL) and introducing two EBIR datasets, \textit{i.e.}, N-UKbench and EC180. For the framework, we first encode the neuromorphic events into understandable tensors, \textit{i.e.}, event image. Then we propose a deep adversarial learning architecture to align the feature distributions of the two modalities of data.
Concretely when an event image/color image is embedded in the joint space, the embedding vector is fed into a modality discriminator which acts as an adversary of feature generators. The feature generators learn modality-invariant features by confusing the discriminator while the discriminator aims at maximizing its ability to identify modalities. For the datasets, N-UKbench is converted from existed frame-based instance retrieval dataset UKbench~\cite{1641018}, in which 2550 event streams and 7650 color images are evenly distributed in 2550 instances. And to further verify the effectiveness of proposed method for real-world data, EC180 is collected by finding different objects to record, in which 180 event streams and 900 color images are evenly distributed in 180 instances.

To sum up, the main contributions of this paper are:
\begin{itemize}
	\item The formulation of a new Event-Based Image Retrieval (EBIR) problem to study the cross-modal matching, requiring that retrieve images of a particular object given an event stream as query.
	\item A new method called neuromorphic Events-Color image Feature Learning (ECFL) for performing EBIR, which generates representations that are discriminating among instances and invariant with respect to modalities to capture correlations across neuromorphic events and color images.
	\item The collection of N-UKbench and EC180 dataset specifically for EBIR to advance the problem. Dataset website: (double-blind review).
\end{itemize}
\section{Related Work}

\subsection{Cross-modal Retrieval}
Data acquired from heterogeneous sensors constitutes our digital modern life. As multi-modal data grows rapidly, cross-modal retrieval has drawn great attention due to its widespread application prospects, such as multi-sensor information fusion, object recognition and scene matching. It aims to take one modality of data to retrieve relevant data of another modality. Till now, multi-modal retrieval has been widely studied, including retrieving infrared images with visible images~\cite{li2018cross-domain}, retrieving videos with texts~\cite{7410872}, retrieving photos with sketches~\cite{wang2016deep}, etc. The major concerns are how to perform the cross-modal correlation modeling to learn common representations for various modalities of data so that the similarity between different modalities can be measured.

There is a significant amount of work in learning a common subspace shared by
different modalities of data. Most studies extract the common features which are
robust in multi-modalities of data and achieve multi-modal matching by comparing the similarities between the features. Edge features are widely used since there is usually a certain relationship between the edges in different modalities of data~\cite{radenovic2019fine-tuning}. Besides, some existed descriptors developed in single modality like SIFT~\cite{lowe2004distinctive} are optimized and improved as feature representation methods for cross-modalities. Although there are several ways to extract features for neuromorphic events and color images, the methods developed on neuromorphic events is not applicable to color images and vice versa. Another line is to project the features of different modalities to a common feature subspace, in which the similarities are measured. Due to the recent progress of deep learning,  the Convolutional Neural Networks (CNN) acting as high-level feature extractors can generate modality-invariant representations, which can be categorized  into two types: the classification-based network and the verification-based network~\cite{zheng2018sift}. The former is trained to classify architectures into pre-trained categories and the learned embedding, \textit{e.g.}, FC7 in Alexnet, is usually employed as cross-modality feature. The latter may use a siamese network and employ the contrastive loss to learn common representations. 

Although cross-modal matching among data of multi-sensors is a hot spot in the field of computer vision, to our best knowledge, there is no studies about cross-modal matching between neuromorphic events and color images yet due to the novelty of event-based vision field. Owing to the unconventional output of event camera and its inconsistent encoding content with conventional camera, existing cross-modal retrieval methods fail to tackle the EBIR task. In this work, we embed the adversarial training strategy into the process of representation learning to bridge the gap between neuromorphic events and color images.

\subsection{Event-based Datasets}
One of the key barriers towards EBIR task is the scarcity of well-annotated datasets. Prior to our work, several datasets for development and evaluation of various event-based methods have already been collected~\cite{gallego2019event}. For example, N-Caltech101 is an event-stream dataset for object recognition which is converted from existing computer vision static image dataset Caltech101~\cite{feifei2007learning} using an actuated pan-tilt camera platform~\cite{ncaltech101}; DET is a high-resolution dynamic vision sensor dataset for lane extraction~\cite{det}; and the MVSEC dataset is used to perform a variety of 3D perception tasks, such as feature tracking, visual odometry, and stereo depth estimation~\cite{mvsec}. However, none of them are designed for the EBIR task.

Since a limited number of datasets are available in the community of neuromorphic vision, it is very difficult to develop a neuromorphic vision dataset by crawling data directly from the web like developing a computer vision dataset. Inspired by the methods of creating datasets in~\cite{ncaltech101, li2017cifar10-dvs:}, which convert existing computer vision static image datasets into neuromorphic vision datasets, we convert the popular frame-based instance retrieval dataset UKbench~\cite{1641018} to a dataset suitable for EBIR task, named as ``N-UKbench''. Moreover, the neuromorphic events converted from static images may be different from that obtained by photographing the real objects. To provide access to evaluating algorithms applied to real scenes, we further provide EC180 dataset, which is collected by finding different objects to record. Such two datasets are important to exploit methods of cross-modal matching between neuromorphic events and color images.

\section{Event Representation}
Event camera works radically different from the frame-based camera. Pixels in an event camera work asynchronously to respond the change of their log photocurrent $\Delta L(I(\mathbf{x}_\mathbf{i},t_i))$. Here $I(\mathbf{x}_\mathbf{i},t_i)$ is the photocurrent (brightness) of the $i^{th}$ event $ev_i$ at time $t_i$ that locates at $\mathbf{x}_\mathbf{i}$, and $L(I(\mathbf{x}_\mathbf{i},t_i)) = log(I(\mathbf{x}_\mathbf{i},t_i))$ indicates the logarithmic operation of the photocurrent. When the brightness change since last event at the pixel $\mathbf{x}_\mathbf{i}$ with duration $\Delta t_i$:
\begin{equation}
\Delta L(I(\mathbf{x}_\mathbf{i},t_i)) = L(I(\mathbf{x}_\mathbf{i},t_i)) - L(I(\mathbf{x}_\mathbf{i},t_i - \Delta t_i)), 
\end{equation}
reaches threshold $\pm C$ ($C > 0$), the event camera will generate a new event with polarity $p_{i} \in\{+1,-1\}$ indicating the increase (\textit{ON} events) or decrease (\textit{OFF} events) of brightness. Then, a stream of events can be defined as:
\begin{equation}
ev_i = [\mathbf{x}_\mathbf{i},t_i,p_i]^T,  \quad i \in \mathbb{N}
\end{equation}
where $\mathbf{x}_\mathbf{i}=[x_i,y_i]^T$, $t_i$ and $p_i$ is the location, trigging time and polarity of the $i^{th}$ event $ev_i$ in the stream.

The output of event camera is a discrete spatial-temporal event stream (as shown in Fig~\ref{fig:mission}), which is not able to be processed by methods developed on the conventional frame-based camera. Thus we convert the event stream into a fix-sized tensor representation. To testify the robustness of proposed model, we adopt three event representation methods including event stacking, time surface, and event frequency.

\subsection{Event Stacking (ES)}
As the most straightforward way to encode an event stream into a tensor, the event stacking has been applied in many tasks including steering prediction~\cite{maqueda2018event}, visual-inertial odometry~\cite{rebecq2017real}, and video generation~\cite{wang2019event}, \textit{etc}. Unlike previous stacking methods, we stack all events ignoring their polarity into one tensor $S(\mathbf{x}),  \mathbf{x} = [x,y]^T$ to prevent the context from being separated into two dispersed part. Thus we can integrate the events within the interval $T_s$ into a tensor using:
\begin{equation}
S(\mathbf{x}) \triangleq \sum_{t_{i} \in T_s} \delta\left(x-x_{i}, y-y_{i}\right),
\end{equation}
where $\delta$ is the Kronecker delta, and $[x_{i}, y_{i}]$ is the pixel position of an event. 

\begin{figure*}[t]
	\centering
	\includegraphics[width=2\columnwidth]{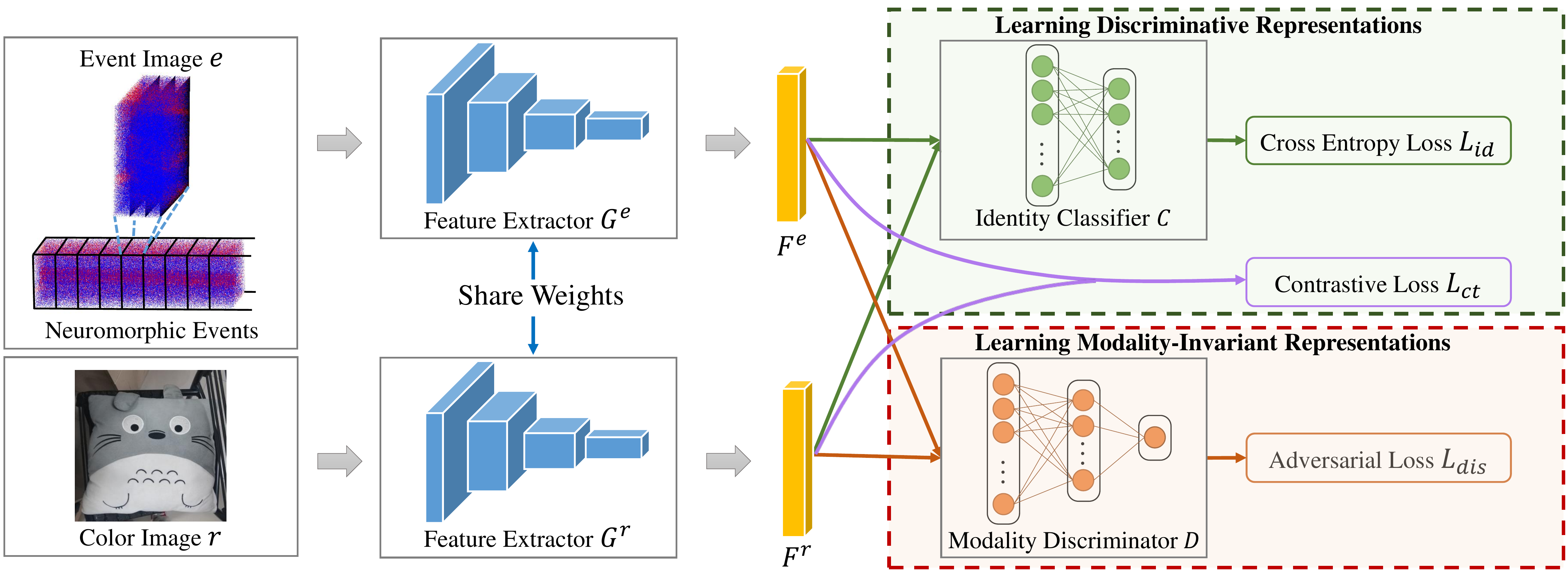} 
	\caption{Architecture of the proposed model. An event image/color image pair $\{e,r\}$ is given as input. Feature extractors $G^e$ and $G^r$ respectively map the event image and color image to $d$-dimensional feature representation $\{F^e,F^r\}$. At training time, the cost function contains two parts. The first part learns features that are distinguishable for different instances, including cross entropy loss and contrastive loss, where the cross entropy loss is calculated by the output of identity classifier $C$. The second part utilizes the adversarial training strategy to learn modality-invariant representations, where a modality discriminator $D$ acts as an adversary. At test time, the discriminator and the classifier are stripped off and the feature generators can map the event image and color image into a common embedding space. }
	\label{fig:model}
\end{figure*}

\subsection{Time Surface (TS)}
The time surface is a spatial-temporal representation of asynchronous event stream. It has been adopted in tasks like object classification~\cite{HOTS} and action recognition~\cite{N-HAR}. To generate a time surface $T_{i}(\mathbf{u})$ of an event $ev_i$, we need to first generate the Surface of Active Events (SAE)~\cite{evflow}:
\begin{equation}
\mathcal{A}_{i}(\mathbf{u}) \triangleq \max _{j \leq i}\left\{t_{j} | \mathbf{x}_{\mathbf{j}}=\left(\mathbf{x}_{\mathbf{i}}+\mathbf{u}\right)\right\},
\end{equation}
where $\mathcal{A}_{i}$ is the SAE around the incoming event $ev_i$ for the pixels in the $(2R+1) \times (2R+1)$ square, centered at the pixel position $\mathbf{x_i} = [x_i,y_i]^T$. And $\mathbf{u} = [u_x,u_y]^T$ is the pixel in that square, $u_{x} \in\{-R, \ldots, R\}$, $u_{y} \in\{-R, \ldots, R\}$. Then we can apply the exponential decay kernel with time constant $\tau_{e}$ to generate a dynamic spatial-temporal time surface $T_{i}(\mathbf{u})$:
\begin{equation}
T_{i}(\mathbf{u}) \triangleq exp({\frac{\mathcal{A}_{i}(\mathbf{u})-t_{i}}{\tau_{e}}})
\end{equation}

\subsection{Event Frequency (EF)}
Besides above representations, we can also analyze events from frequency domain. The event frequency~\cite{miao2019} counts the event occurrence at each pixel within a range as activation frequency to encode the event stream into a tensor. Since the noise in an event camera has relative low frequency, the representation is more friendly to noise tolerance. The event frequency $F(\mathbf{x},n)$ can be generated by using:
\begin{equation}
F(\mathbf{x},n) \triangleq 1 - \frac{2}{exp(n)+1}
\end{equation}
where $n$ is the total number of the activated events at pixel $\mathbf{x} = [x,y]^T$ within a time interval $T_f$. 

\section{Cross-modal feature learning}
\subsection{Overview}
We focus on the task of neuromorphic event and color image bimodal representation learning. We divide the event stream in 90 ms intervals, where each bin is converted to an ``event image". To make full use of the temporal information of the neuromorphic events, we further divide each bin into three consecutive parts. Each part is processed by the event representation method and acts as a channel of the event image. Let $\mathcal{C}$ be the set of all possible instances in a given dataset; $\mathcal{E}=\{(e_i,c_i^e)\}^N_{i=1}, c_i^e\in{\mathcal{C}}$ and $\mathcal{R}=\{(r_i,c_i^r)\}^M_{i=1}, c_i^r\in{\mathcal{C}}$ be the set of event images and color images respectively, where $e_i$ is an event image, $r_i$ is an original color image, $\{c_i^e,c_i^r\}$ are their corresponding instance-level object identity label, and $\{M,N\}$ are the total number of event images and color images respectively. As event images $\mathcal{E}$ and color images $\mathcal{R}$ clearly have different statistical properties and follow unknown distributions, they fail to be directly compared against each other. Our goal is to learn a common subspace in which the similarity between data items from these two modalities (neuromorphic events and color images) can be assessed directly, \textit{i.e.}, learned features have the same or very similar distributions in the two modalities of data. 

Towards this goal, our proposed ECFL method described in Fig.~\ref{fig:model} embeds adversarial training strategy into the process of representation learning, so that the learned features combine discriminativeness and modality-invariance. Specifically, the aim of our framework is to learn two feature extractors $G^e:\mathcal{E}\rightarrow\mathbb{R}^d$ and $G^r:\mathcal{R}\rightarrow\mathbb{R}^d$ which respectively map the event image and color image into a common embedding space. The cost function which guides the learning process to provide the embedding with the desired properties contains two parts. The first part learns features that are distinguishable for different instances, including cross entropy loss and contrastive loss. The cross entropy loss is calculated by the output of identify classifier $C:\mathbb{R}^d\rightarrow\mathcal{C}$, which takes the feature representation as input and predicts the object identity label. It makes sure that the learned embedding space preserves instance-discriminative information. The identify classifier is composed of a softmax layer, whose number of output nodes equals to the number of instances $N_{ins}$. The contrastive loss is calculated directly in the embedding space, which ensures that the distance between the event image and color image is minimized if they denote the same instance and is greater than a margin value if they denote two different instances. The second part utilizes the adversarial training strategy to minimize the distribution distance between data items from two modalities, such that the feature distributions are well aligned in the embedding space, where a modality discriminator $D:\mathbb{R}^d\rightarrow[0,1]$ acts as an adversary. It predicts $0$ for event image and $1$ for color image. We regard a representation as a modality-invariant feature when the modality discriminator fails to discriminate modalities they belong to. The modality discriminator is composed of two FC layers and a sigmoid function.

The architecture of feature extractors $G^e$ and $G^r$ are based on that of ResNet~\cite{he2016deep}, which has enough capacity of representations considering large margin of statistical properties between the event images and color images. The weights of the two feature extractors are shared everywhere. The weight-sharing constraint allows to align the distributions of two modalities from the very beginning of the feature extraction. Besides, there is another strategy: without any weight-sharing. It learns features of two modalities independently and relies entirely on the cost function to guide the alignment of them. But it can preferably deal with the heterogeneous nature of the input. Both strategies are evaluated in our experiment. 

\subsection{Model Learning and Deployment}
The feature extractors learn representations that are invariant to different modalities by constantly trying to outsmart the modality discriminator, which is working to become a better detective. At the same time, they learn representations that are distinguishable for different instances. Therefore, the learning stage contains two steps: (i) Update the modality discriminator $D$ by minimizing the loss of the modality discriminator; (ii) Update the feature extractors $G^e$ and $G^r$ by maximizing the loss of the modality discriminator while minimizing the cross entropy loss and contrastive loss. Let us consider a training batch $\mathcal{D}=\{e_i,r_i\}^n_{i=1}$ including $n$ event image-color image pairs, where $e_i\in\mathcal{E}$ and $r_i\in\mathcal{R}$. The objective of the modality discriminator is to predict the modality of the input given its representation and can be written as:
\begin{equation}
L_{dis}=-\frac{1}{n}\sum_{i=1}^{n}{{\rm log}(D(F_i^e))+{\rm log}(1-D(F_i^r))},
\label{Eq:Ldis}
\end{equation}
where $F_i^e=G^e(e_i)$ and $F_i^r=G^r(r_i)$. The objective of the feature generator is now to defeat the modality discriminator, \textit{i.e.}, the modality discriminator should not be able to predict the modality of the input given its representation. In addition, it has to minimize the cross entropy loss and contrastive loss. The complete objective is then:
\begin{equation}
L=\alpha{L_{id}}+\beta{L_{ct}}-\gamma{L_{dis}},
\label{Eq:L}
\end{equation}
where $L_{id}$ is the cross entropy loss for predicting the object identity label:
\begin{equation}
L_{id} = -\frac{1}{n}\sum_{i=1}^{n}{(\sum_{j=1}^{N_{ins}}{\hat{p_{ij}^e}{\rm log}(p_{ij}^e)}+\sum_{j=1}^{N_{ins}}{\hat{p_{ij}^r}{\rm log}(p_{ij}^r)})}, 
\end{equation}
\begin{equation}
p_{ij}^{\{e,r\}} = \frac{{\rm exp}(C(F_i^{\{e,r\}}))}{ \sum_{k=1}^{N_{ins}}{C(F_i^{\{e,r\}})}},
\end{equation}
$L_{ct}$ is the contrastive loss, whose goal is to make the distance between an event image and a color image of the same instance closer than that the one of two different instances:
\begin{equation}
L_{ct} = \frac{1}{2n}\sum_{i=1}^{n}{sl_i^2+(1-s){\rm max}(m-l_i,0)^2},
\end{equation}
\begin{equation}
l_i=||F_i^e-F_i^r||_2,
\end{equation}
$s=1$ if $e_i$ and $r_i$ denote the same instance else $s=0$ and $m$ is the margin. $\alpha$, $\beta$ and $\gamma$ control the contribution of the three items respectively. Suppose $\theta_G$, $\theta_C$ and $\theta_D$ are the parameters of feature generators, identity classifier and modality discriminator, respectively. We apply an alternating gradient update scheme similar to the one described in~\cite{goodfellow2014generative} to seek  $\theta_G$, $\theta_C$ and $\theta_D$, as shown in Algorithm~\ref{algorithm:training}. 
\begin{algorithm}[th]
	\caption{Optimizing our proposed model}
	\label{algorithm:training}
	\KwData{training set $\mathcal{E}$ and $\mathcal{R}$, margin $m$, weighting factors $\alpha$, $\beta$ and $\gamma$.}
	\Repeat{Convergence or max training iterations}{
		Get a random mini-batch $\mathcal{D}=\{e_i,r_i\}^n_{i=1}$\;
		\nl Update $D$, with $G^e$ and $G^r$ fixed:\\
		\quad Compute $L_{dis}\leftarrow{Eq. \ref{Eq:Ldis}}$\;
		\quad $\theta_D \leftarrow \theta_D - \eta\bigtriangledown_{\theta_D} L_{dis}$\;
		
		\nl Update $G^e$ and $G^r$ as well as $C$, with $D$ fixed:\\
		\quad Compute $L\leftarrow{Eq. \ref{Eq:L}}$\;
		\quad $[\theta_G, \theta_C] \leftarrow [\theta_D,\theta_C] - \eta\bigtriangledown_{\theta_D,\theta_C} L$\;
	}
\end{algorithm}

After learning, the modality discriminator and the identity classifier are stripped off during testing. We use the feature generators to map the event images and color images to the d-dimensional feature representations. Thus the similarity of neuromorphic events and color images can be simply measured by Euclidean distance. Color images in database are ranked according to the similarity, and the ones with similarity at the top are set as matching results. 

\section{N-UKbench and EC180 Dataset}

Datasets are fundamental tools to facilitate adoption of event-driven technology and advance its research~\cite{gallego2019event}. A good dataset should meet the requirements of algorithm prototyping, deep learning and algorithm benchmarking. In this paper, we contribute N-UKbench and EC180 dataset specifically for EBIR task.

N-UKbench is converted from an existed instance retrieval dataset UKbench~\cite{1641018}. UKbench consists of 2550 instances, where each instance includes 4 color images under various angles, illuminations, translations, etc. To make the dataset applicable to our EBIR task, we convert 1 image of each instance to 1 event stream and the rest of 3 color images depicting the same instance are remained. We use the method in~\cite{li2017cifar10-dvs:} to convert the static image to event stream, \textit{i.e.}, using the CeleX-V~\cite{celex5} event camera to record the screen on which image is moving. the CeleX-V~\cite{celex5} is the current highest resolution event camera, whose resolution is $1280\times800$ pixels.

\begin{figure}[!t]
	\vfill
	\subfloat[Recording Setup]{%
		\includegraphics[clip,width=0.41\columnwidth]{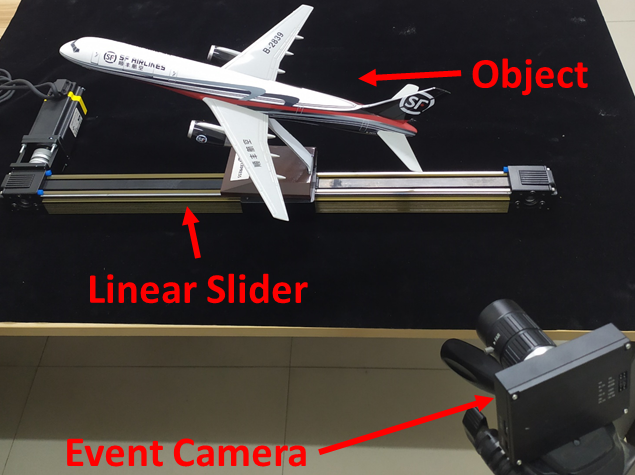}%
		\label{fig:recording}
	}
	\hfill
	\subfloat[Example data in EC180]{%
		\includegraphics[clip,width=0.55\columnwidth]{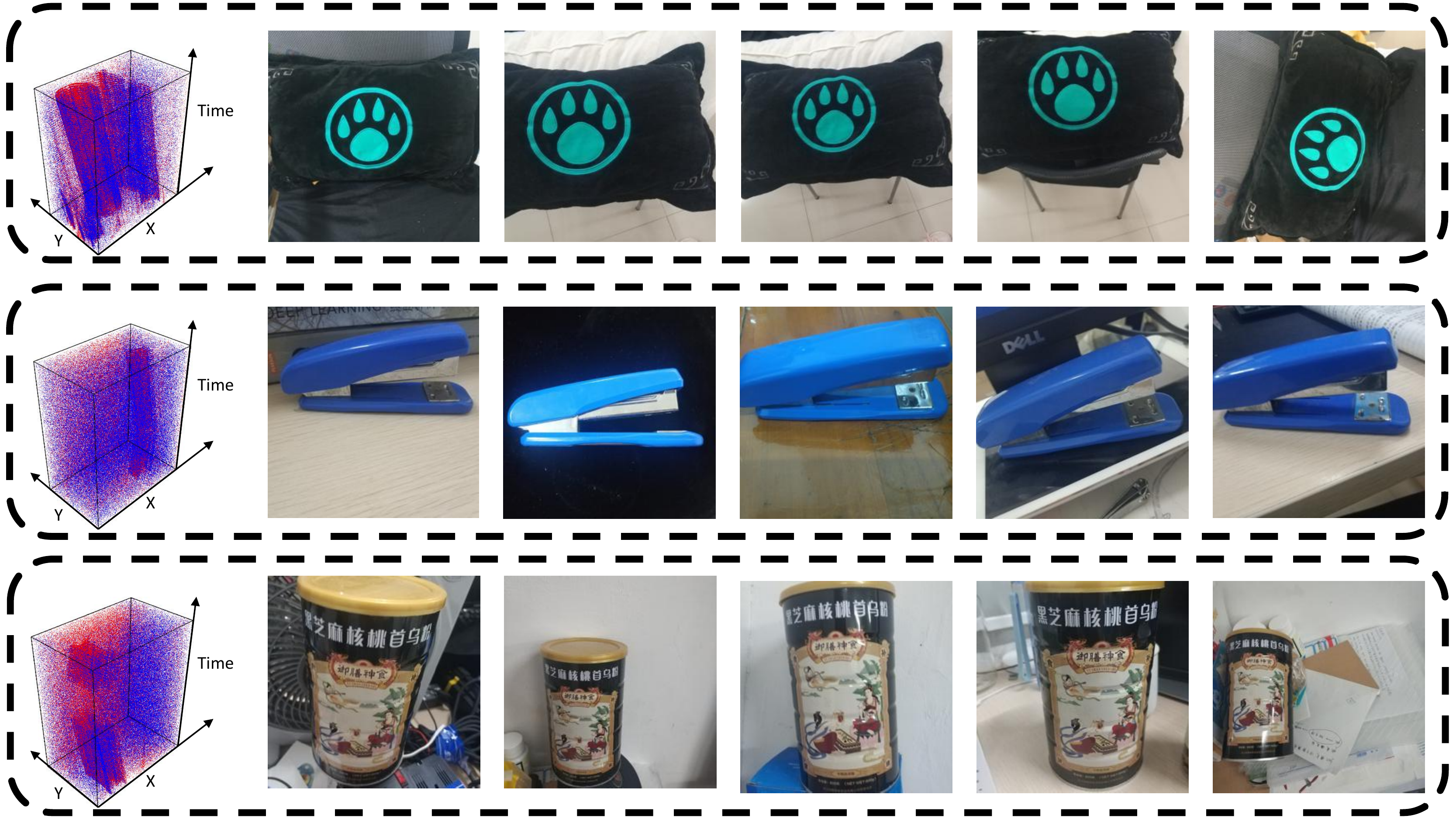}%
		\label{fig:examples}
	}
	\caption{Recording setup and samples in EC180 dataset. (a) is the recording setup of EC180 dataset. We use a linear slider to generate relative motion between object and event camera. (b) gives three samples in EC180 dataset. }
	\label{fig:dataset}
\end{figure}

EC180 dataset is collected by photographing real objects. There are overall 180 instances in it. Since neuromorphic events can only be generated when there is relative motion between the event camera and the object, most recordings in EC180 are conducted in the setup shown in Fig.~\ref{fig:recording}. The linear slider will drive the object to move so that the event camera can ``see" the object. There are also several objects that are recorded by moving the event camera. The color images in EC180 dataset are captured with different background and perspective. Fig.~\ref{fig:examples} gives three samples in our EC180 dataset, where each sample contains 1 event stream and 5 color images.

\section{Experiments}
\subsection{Experiment Settings}
We use the Pytorch to train our models. The modality discriminator is trained with Adam~\cite{kingma2014adam}, using a learning rate of 0.002 and $[\beta1,\beta2]=[0.5,0.99]$. The feature generators and the identity classifier are trained with SGD~\cite{bottou2010large}, using a learning rate of 0.001 and a momentum of 0.9. The weighting factors $\alpha$, $\beta$ and $\gamma$ are set to 1, 0.01, 0.01. The maximum epoch of training iterations is set to 20.

We randomly select 2040 instances from N-UKbench as training set and the remaining 510 instances are for evaluation. We rotate each event image/color image to 9 angles ranging from $-45\degree$ to $45\degree$ and double the numbers by flipping them horizontally. All event images and color images are resized to the same size of $224\times224$ pixels.  

\subsection{Performance Analysis}

\begin{figure}[!t]
	\centering
	\includegraphics[width=1\columnwidth]{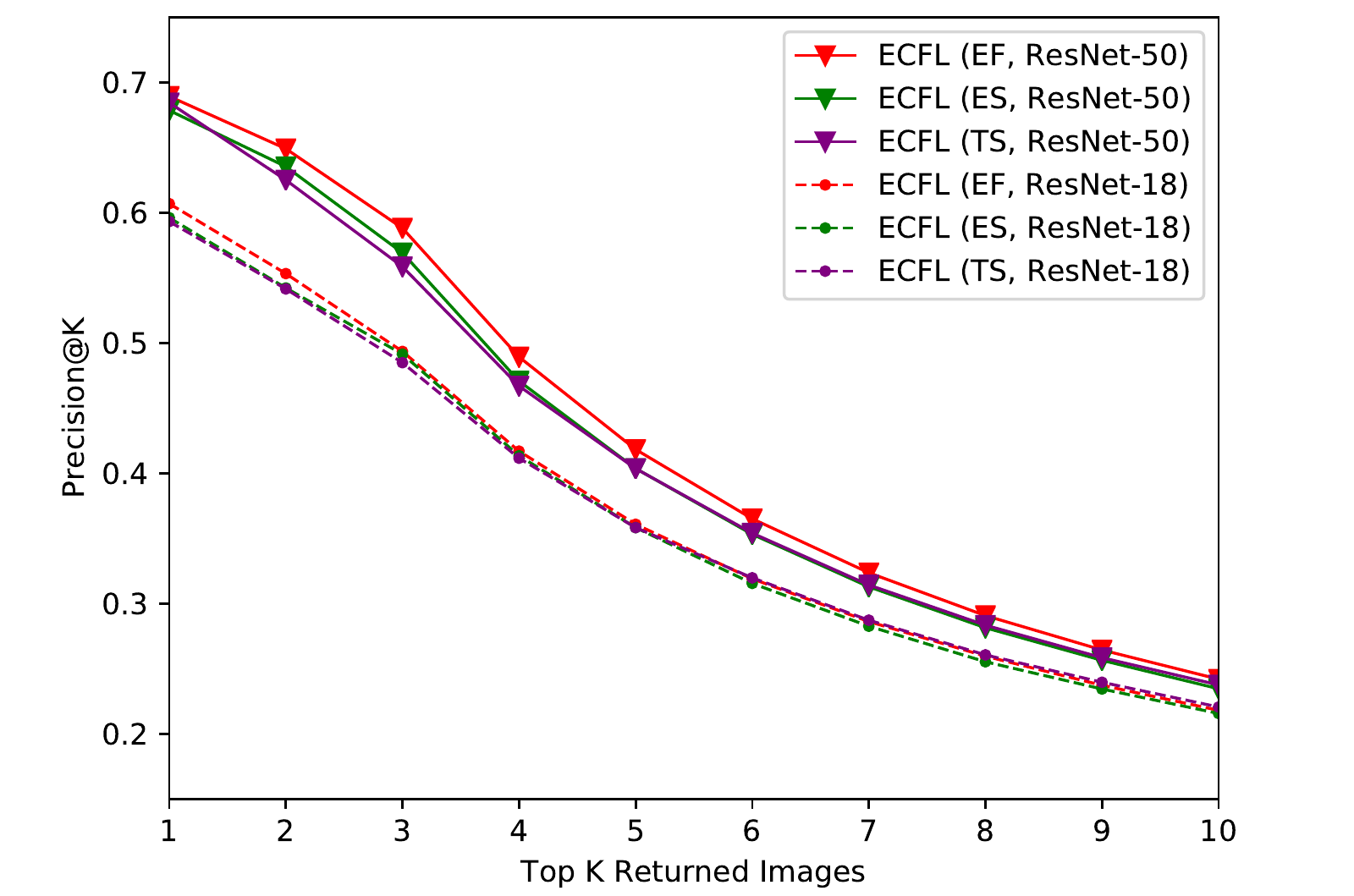} 
	\caption{Top $K$ precision of ECFL with different event representations.}
	\label{fig:pr}
\end{figure}

\begin{table}[!t]
	\caption{Retrieval performance of SIFT, LSS and ECFL with different event representations}
	\small
	\resizebox{\linewidth}{!}{
		\begin{tabular}{l|l|l|l}
			\hline
			&  mAP    &  acc.@1 &  acc.@3  \\ 
			\hline
			SIFT (ES)               &   0.0491         &      0.0373      &       0.0346     \\
			SIFT (TS)               &   0.0559         &      0.0554      &       0.0415     \\
			SIFT (EF)               &   0.0550         &      0.0426      &       0.0364     \\
			LSS (ES)                &   0.0394         &      0.0289      &       0.0230     \\
			LSS (TS)                &   0.0310         &      0.0157      &       0.0201     \\
			LSS (EF)                &   0.0247         &      0.0172      &       0.0163     \\
			ECFL (ES, ResNet-18)    &   0.5676         &      0.5961      &       0.4913     \\
			ECFL (TS, ResNet-18)    &   0.5683         &      0.5931      &       0.4848     \\
			ECFL (EF, ResNet-18)    &   0.5724         &      0.6069      &       0.4935     \\
			ECFL (ES, ResNet-50)    &   0.6463         &      0.6784      &       0.5691     \\
			ECFL (TS, ResNet-50)    &   0.6419         &      0.6843      &       0.5585     \\
			ECFL (EF, ResNet-50)    &  \textbf{0.6651} &  \textbf{0.6892} &  \textbf{0.5881} \\
			\hline 
		\end{tabular}
	}
	\label{tab:representation}
\end{table}

\begin{figure*}[!t]
	\vfill
	\subfloat[Samples in N-UKbench]{%
		\includegraphics[clip,width=1\columnwidth]{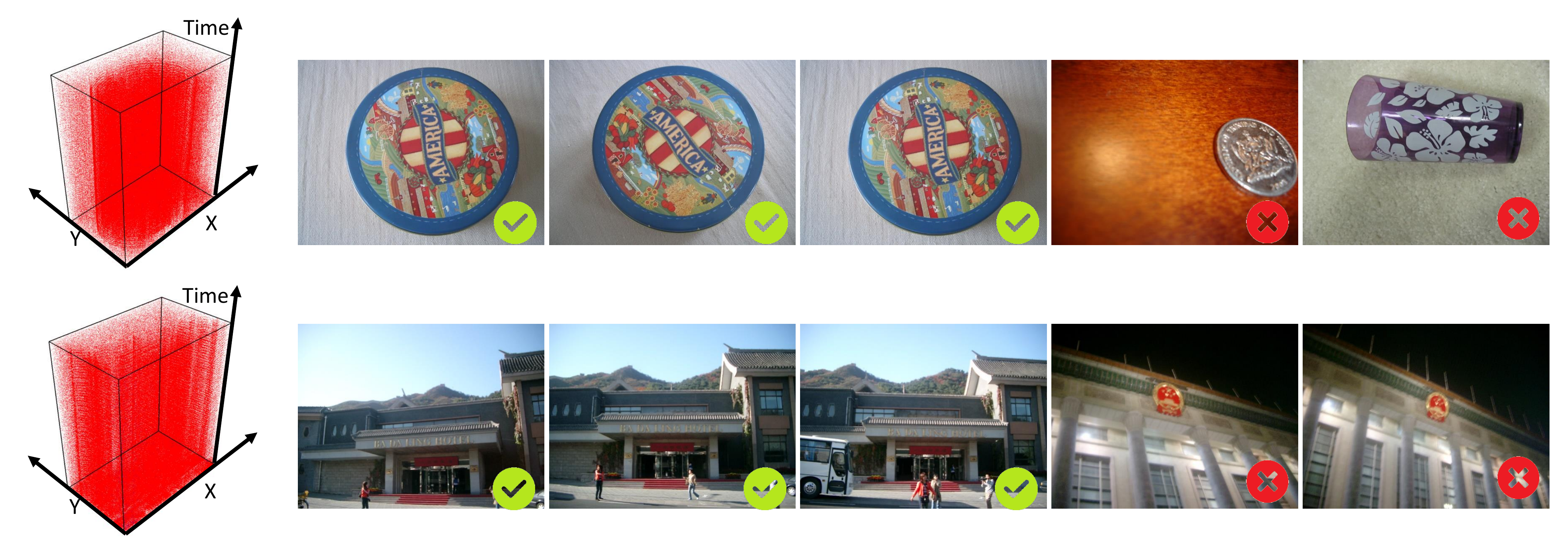}%
		\label{fig:result1}
	}
	\hfill
	\subfloat[Samples in EC180]{%
		\includegraphics[clip,width=1\columnwidth]{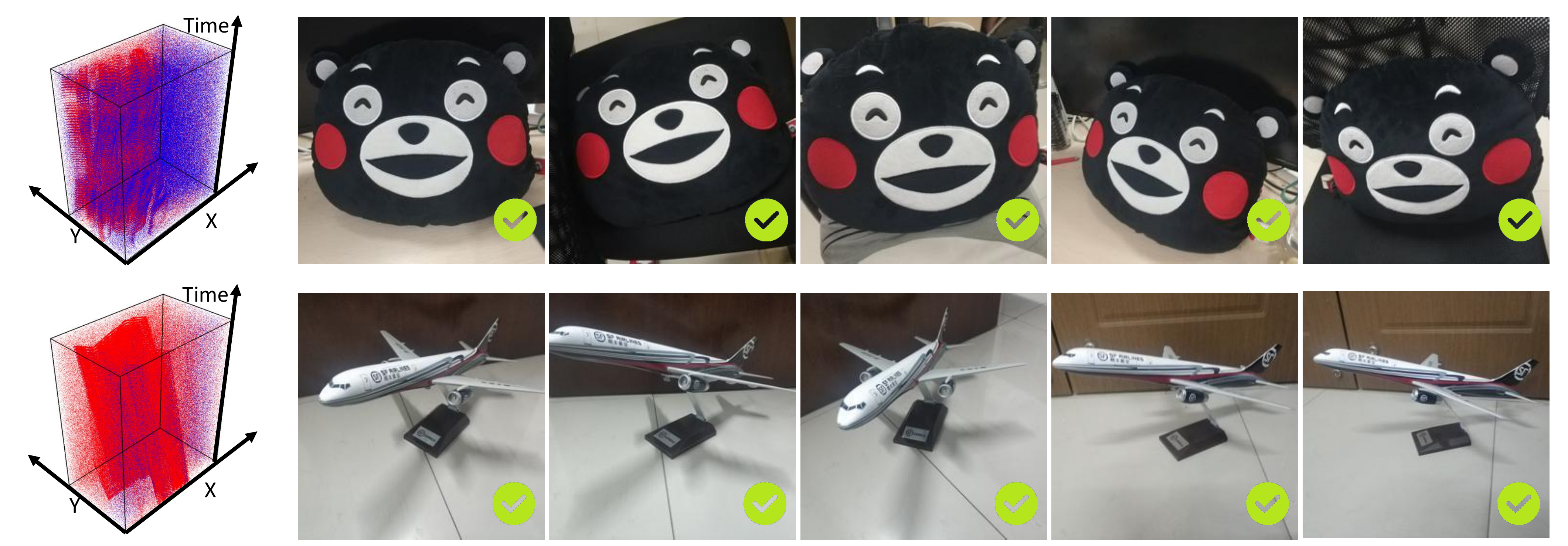}%
		\label{fig:result2}
	}
	\caption{Qualitative evaluation of our proposed model. For each sample, the top 5 retrieved results with different event representations are shown in each row. The green tick in the lower right corner of the result represents that it is a true match while the red cross represents a false match.}
	\label{fig:results}
\end{figure*}

The purpose of this paper is to retrieve images of a particular object given an event stream as query. To the best of our knowledge, there are no studies about cross-modal matching between neuromorphic events and color images. Nevertheless, we still compare ECFL with Scale Invariant Feature Transform (SIFT)~\cite{lowe2004distinctive} and Local Self-Similarity (LSS)~\cite{4270223} because the event cluster in each stream forms the shape of the object. SIFT is the most representative in describing the structure information. We employ SIFT in combination with spatial pyramid~\cite{1641019} to describe both event images and color images and measure the similarity by histogram intersection. LSS ia able to capture shape features of a large image region and it is suitable for images with complex radiometric differences, which is widely adopted to multi-modal image matching, such as optical-to-SAR image matching~\cite{li2018rift} and sketch-to-image matching~\cite{4270223}. We use LSS with BoW~\cite{1238663} to describe event images and color images and measure the similarity by Euclidean distance. Tab.~\ref{tab:representation} shows the mean average precision (mAP) and the precision of top K retrieval results (acc.@K, K=1,3) of SIFT, LSS and ECFL while employing different event representations. Fig.~\ref{fig:pr} shows the precision of top K retrieval results of ECFL and Fig.~\ref{fig:result1} shows examples of retrieval results of ECFL using event frequency as representation. For each sample, the top 5 retrieved results are shown in each row. 

It is shown that our proposed ECFL method (for each event representation) performs the best. It is able to bridge the large gap between neuromorphic events and color images and return the true matches. SIFT and LSS descriptors performs poorly for the novel EBIR task. Because the event image is different from the color image, it is produced by brightness changes, which results in the poor distribution consistency of their features. As shown in Fig.~\ref{fig:pr}, the retrieval performance of the three event representations is very close. Among them, the event frequency performs slightly better. Comparing the other two event representation methods, the event frequency can significantly filter out the noise caused by the sensor since the occurrence frequency of noise at a particular pixel is low. Our proposed ECFL method, including ResNet-18 and ResNet-50 as backbone, produce a significant performance improvement. ResNet-50 architecture performs better than ResNet-18. It is expected that deeper architecture have more parameters, and can therefore better cope with the inconsistent distribution of features.

\begin{figure*}[!t]
	\vfill
	\subfloat[Sample in N-UKbench]{%
		\includegraphics[clip,width=1\columnwidth]{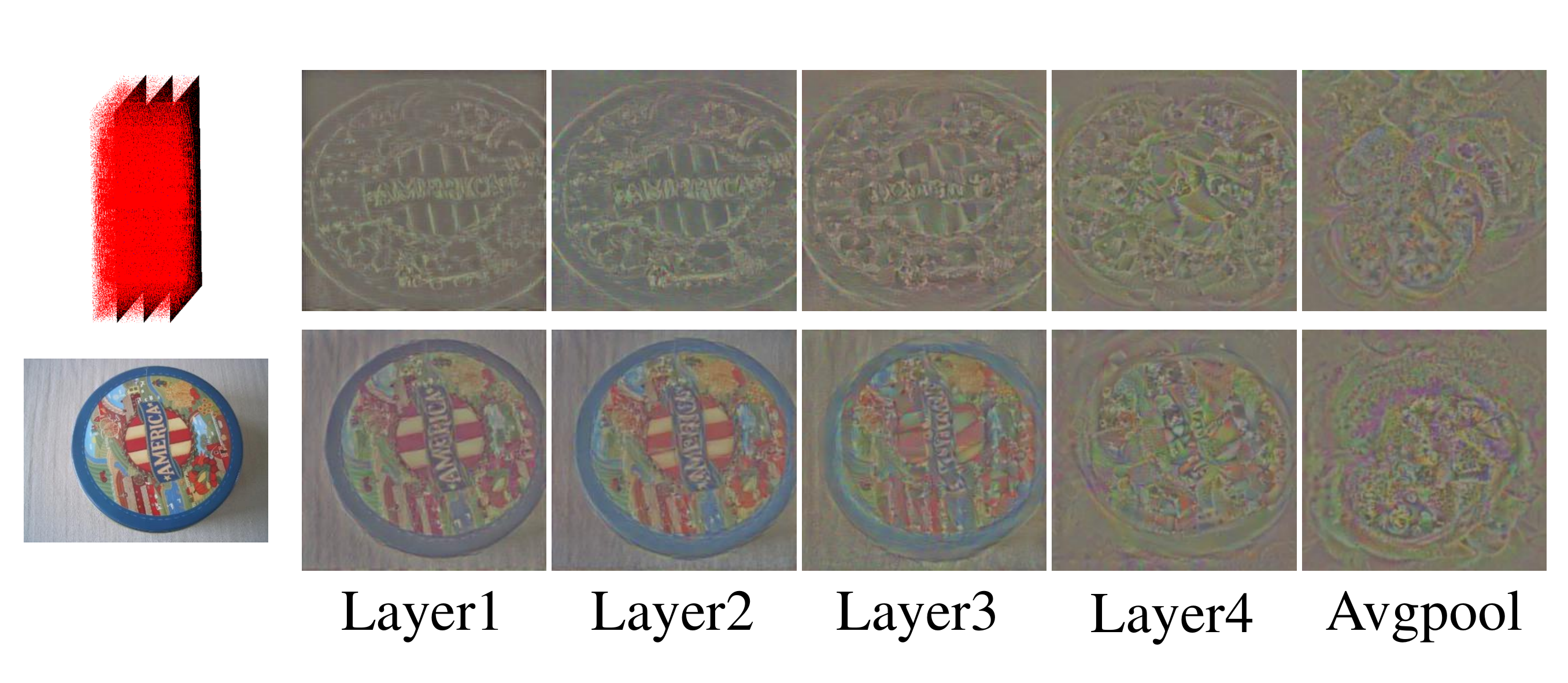}%
		\label{fig:visualization1}
	}
	\hfill
	\subfloat[Sample in EC180]{%
		\includegraphics[clip,width=1\columnwidth]{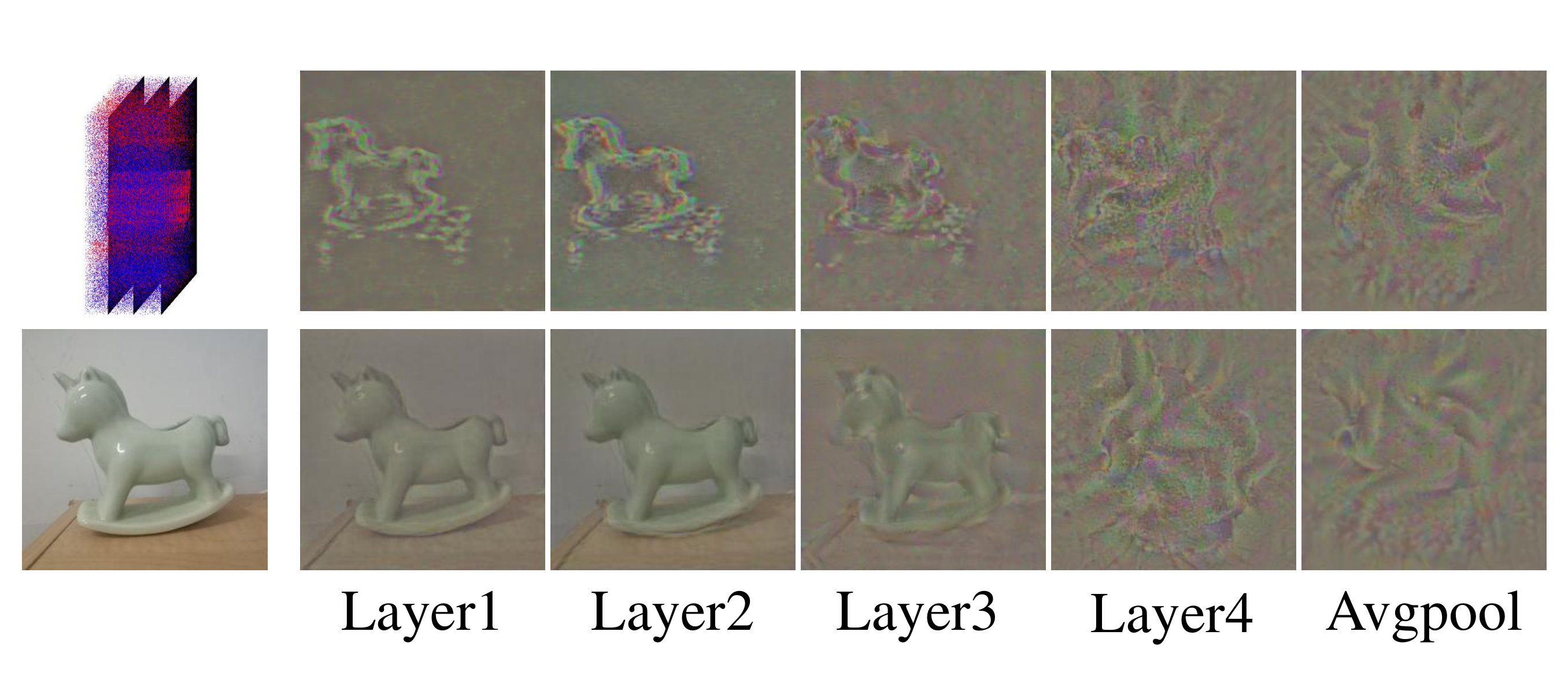}%
		\label{fig:visualization2}
	}
	\caption{CNN reconstruction of the input data from layer 1 to 4, and the avgpool layer. All the generated images are reconstructed by the features extracted from every corresponding layer's last layer.}
	\label{fig:visualization}
\end{figure*}

\begin{table}[!t]
	\caption{Quantitative evaluation with the strategy of No Weight-Sharing (NWS)}
	\small
	\begin{tabular}{p{3.78cm}|p{0.9cm}|p{0.9cm}|p{0.9cm}}
		\hline 
		&   mAP &  acc.@1 & acc.@3   \\ 
		\hline
		ECFL (ES, ResNet-50, NWS)       & 0.6284 & 0.6456 & 0.5539 \\
		ECFL (TS, ResNet-50, NWS)       & 0.6293 & 0.6676 & 0.5462 \\
		ECFL (EF, ResNet-50, NWS)       & 0.6303 & 0.6480 & 0.5564 \\
		\hline 
	\end{tabular}	
	\label{tab:weight-sharing}
\end{table}

\subsection{Impact of Different Weight-Sharing Strategies}
It has been verified that the strategy of all weight-sharing is better suited for situations where two inputs are comparatively similar, \textit{e.g.}, image and sketch~\cite{7780462}, while the strategy of no weight-sharing is better suited for situations where two inputs differ somewhat, \textit{e.g.}, image and text~\cite{7780462}. So, what about the neuromorphic events and color images? In this experiment, we compare two weight-sharing strategies described above. Tab.~\ref{tab:weight-sharing} shows the quantitative evaluation results of our proposed model with the strategy of no weight-sharing. It is easy to see that our ECFL method with all weight-sharing performs better than without any weight-sharing in aligning the feature distributions of two modalities of data. We consider that although the output of event camera is sparse and non-uniform spatiotemporal event signal, it is also a type of vision sensor. After converting the event stream into event images, the two modalities of data can be regarded as similar.

\begin{table}[!t]
	\caption{Quantitative evaluation with No use of Temporal Component (NTC)}
	\small
	\begin{tabular}{p{3.78cm}|p{0.9cm}|p{0.9cm}|p{0.9cm}}
		\hline 
		&  mAP &  acc.1 &  acc.@3  \\ 
		\hline
		ECFL (ES, ResNet-50, NTC)    & 0.5501 & 0.5686 & 0.4796 \\
		ECFL (TS, ResNet-50, NTC)    & 0.5762 & 0.6029 & 0.4931 \\
		ECFL (EF, ResNet-50, NTC)    & 0.5727 & 0.5775 & 0.4917 \\
		\hline 
	\end{tabular}
	\label{tab:temporal}
\end{table}
\subsection{Impact of temporal component}
To make full use of the temporal component of the neuromorphic events, we propose a simple but effective approach to model the temporal component of the neuromorphic events, \textit{i.e.}, discretizing each bin into three consecutive parts and treating these parts as different channels in the first convolutional layer. To examine whether temporal component helps the improvement of retrieval performance, in this experiment, we show the quantitative evaluation results of our model with no use of the temporal component. We simply convert each bin into an event image and feed it into the network. Tab.~\ref{tab:temporal} presents the results. We can clearly see that the performance of discarding temporal information is poorer than that of preserving temporal information. We consider supplementary details can be observed along the time axis. By discretizing each fixed time interval into consecutive parts, we can preserve more details and avoid them being offset by each other.

\subsection{Impact of Adversarial Learning}
Here, we investigate the contribution of adversarial learning. In our proposed ECFL model, we use a modality discriminator as an adversary of feature generators. We consider that the feature generators learn modality-invariant features by confusing the discriminator while the discriminator aims at maximizing its ability to identify modalities. In order to evaluate the contribution of adversarial learning, we remove the modality discriminator and train the feature extractors and identity classifier in one step. Tab.~\ref{tab:adversarial} shows the quantitative evaluation results of our model with no adversarial learning. Obviously, for each event representation method, when the modality discriminator is removed, the retrieval performance is degraded. Therefore, adversarial learning can narrow the modality gap between neuromorphic events and color images effectively.

\begin{table}[!t]
	\caption{Quantitative evaluation with No Adversarial Learning (NAL)}
	\small
	\begin{tabular}{p{3.69cm}|p{0.9cm}|p{0.9cm}|p{0.9cm}}
		\hline 
		&  mAP &  acc.1 &  acc.@3  \\ 
		\hline
		ECFL (ES, ResNet-50, NAL)    & 0.6159 & 0.6574 & 0.5538 \\
		ECFL (TS, ResNet-50, NAL)    & 0.6243 & 0.6627 & 0.5482 \\
		ECFL (EF, ResNet-50, NAL)    & 0.6329 & 0.6696 & 0.5510 \\
		\hline 
	\end{tabular}
	\label{tab:adversarial}
\end{table}	

\subsection{Experimental Results on Real Data}
To verify the performance of our algorithm in real scenes, we also evaluate the proposed ECFL method on EC180 dataset. Considering the subtle differences between the neuromorphic events converted from the static image and the neuromorphic events recorded from the real object, our model may suffer performance degradation when applied to real scenes. To avoid it, we take the model trained on the converted N-UKbench dataset as the preliminary model and use the real EC180 dataset to fine-tune it. We randomly select 144 instances in EC180 as training set and the remaining 36 instances are as test set.  Tab.~\ref{tab:ec180} shows the quantitative results after fine-tuning  and Fig.~\ref{fig:result2} shows examples of retrieval results using event frequency as representation.  We can find that our model performs well for real scenes. The retrieval performance is somewhat worse than that on converted N-UKbench dataset. Because the color images in EC180 contain significant amounts of backgrounds and are captured from perspectives with large variants.

\begin{table}[!t]
	\caption{Retrieval performance of EC180}
	\small
	\begin{tabular}{p{3.69cm}|p{0.9cm}|p{0.9cm}|p{0.9cm}}
		\hline
		&  mAP    &  acc.@1 &  acc.@3  \\ 
		\hline
		ECFL (ES, ResNet-50)    &      0.5974      &    0.6111        &          0.5611  \\
		ECFL (TS, ResNet-50)    &      0.5960      &    0.6167        &          0.5463  \\
		ECFL (EF, ResNet-50)    &      0.6102      &    0.6389        &          0.5556  \\
		\hline 
	\end{tabular}
	\label{tab:ec180}
\end{table}

\subsection{Feature Visualization}
Fig.~\ref{fig:visualization} shows the visualized results of different layers in feature generators. The visualization is achieved using the method in~\cite{mahendran2015understanding}, which can invert the learned representation to reconstruct the image to analyze the visual information contained in it. The retrieval performance is often not determined by all the information of the input, but by the main target area. As the layer goes deeper, for the reconstruction of color images, the cluttered background is eliminated and the spatial structure of object gradually changes. It is proved that our model is able to extract the common features of neuromorphic events and color images at avgpool layer. 

\section{Conclusion}
In this paper, we propose the Event-Based Image Retrieval (EBIR) task, which is a novel problem in event-based vision field. Along with the EBIR task, the neuromorphic Events-Color image Feature Learning (ECFL) method has been proposed to address it. We embed the adversarial training strategy into the process of representation learning, so that the learned features combine discriminativeness and modality-invariance. To give an evaluation to our model and stimulate the EBIR task, two event-based color image retrieval datasets, \textit{i.e.}, N-Ukbench and EC180, are collected and will be publicly available soon. Experimental results on our datasets show our model is capable to learn the cross-model representation of neuromorphic events and color images.

As an interesting direction for future work, we plan to extend this work to the case of more than two
sensors. There are many possible combinations can be studied, such as event camera, RGB camera, infrared  camera and LiDAR. Besides that, how to conduct single-modality-based neuromorphic event retrieval is also a direction that deserves researchers' attention.

{\small
\bibliographystyle{ieee_fullname}
\bibliography{egbib}
}

\end{document}